# Constraint Propagation with Imprecise Conditional Probabilities

Stéphane AMARGER    Didier DUBOIS    Henri PRADE

Institut de Recherche en Informatique de Toulouse (I.R.I.T.) - Université Paul Sabatier
118 route de Narbonne
31062 TOULOUSE Cedex - FRANCE
email: [amarger, dubois, prade]@irit.fr

## Abstract

An approach to reasoning with default rules where the proportion of exceptions, or more generally the probability of encountering an exception, can be at least roughly assessed is presented. It is based on local uncertainty propagation rules which provide the best bracketing of a conditional probability of interest from the knowledge of the bracketing of some other conditional probabilities. A procedure that uses two such propagation rules repeatedly is proposed in order to estimate any simple conditional probability of interest from the available knowledge. The iterative procedure, that does not require independence assumptions, looks promising with respect to the linear programming method. Improved bounds for conditional probabilities are given when independence assumptions hold.

## 1 INTRODUCTION

In commonsense reasoning it is very usual to manipulate rules with exceptions. One of the most important cases of such rules consists in default statements containing explicit or implicit numerical quantifiers. Even when they are explicit these quantifiers may be only vaguely stated as for instance in the proposition "most students are young" ; see (Zadeh, 1985). The numerical approach interprets the linguistic term "most" in this example as an ill-defined numerical quantifier expressing the proportion of young people among students, in a certain context (for simplicity we assume here that 'young' has a clear-cut meaning and is not viewed as a fuzzy predicate ; see (Dubois and Prade, 1988) for preliminary results on the handling of fuzzy predicates in this framework). More generally, we may have some imprecise statement about the value of the probability of not encountering an exception, i.e. in our example, the conditional probability P(young | student) is bounded from below by some number in [0,1]. This type of incomplete statistical information is considered by Kyburg (1974) as a large part of our commonsense knowledge. P(young | student) is often also regarded as a degree of certainty that a student taken at random is indeed young.

Different kinds of treatment can be imagined for rules of the kind "if A then B with probability P(B | A)." This can be illustrated considering the above rule and another which can be chained with it, namely, "if B then C with probability P(C | B)." Applying Bayes rule we have P(C | A) ≥ P(B ∩ C | A) = P(C | A ∩ B)·P(B | A) (here we use the same symbol '∩' for denoting the conjunction of propositions or the intersection of the classes of items which satisfy the propositions). Then assuming irrelevance of A with respect to C in the context B, namely assuming here that P(C | A ∩ B) = P(C | B), we obtain the lower bound P(C | B)·P(B | A) for P(C | A). But without this kind of assumption, as soon as P(C | B) ≠ 1, nothing can be said about the value of P(C | A) which can take any value in the interval [0,1]. Indeed nothing forbids to have A ∩ C = ∅ (leading to P(C | A) = 0) as well as A ⊆ C (leading to P(C | A) = 1) for instance. Interestingly enough if we add some information about P(A | B) we may obtain non-trivial bounds for P(C | A) just from bounds on P(B | A), P(A | B) and P(C | B). Then more generally we may choose either i) to exploit the available knowledge on conditional probabilities for computing the best possible upper and lower bounds for some other conditional probabilities of interest, or ii) to take advantage of independence assumptions (which are perhaps hard to check) and prior probabilities for computing probability estimates (Pearl, 1988). The first approach may give no informative result, but when results are informative, they are very strong. On the contrary, the Bayesian approach always gives informative results, but these results can always be questioned by the arrival of new pieces of information. In this paper we investigate the first approach in detail.

Formally, let X be a set of objects, A and B be two subsets of X, and $Q_B^A$ be a subset of values (which may reduce to a single value) expressing what is known about the proportion of A's which are B's. $Q_B^A$ is a subinterval of the unit interval [0,1], corresponding to the default rule "$Q_B^A$ A's are B's." This knowledge is understood as a constraint acting on the cardinality of B relative to A, i.e.:

$$\frac{|A \cap B|}{|A|} \in Q_B^A \subseteq [0,1]$$

where |A| is the cardinality of the subset A. More generally, it is equivalent to a piece of information of the

form P(B | A) ∈ [P_*(B | A), P*(B | A)] where only the two bounds P_*(B | A) and P*(B | A) are known. Indeed relative cardinality is a particular case of conditional probability, where the underlying distribution is uniformly distributed over X. Thus proportions and probabilities obey the same mathematical laws and we shall use them in an interchangeable way in the following.

We use a network representation, as for instance the one on Figure 1, where the two directed edges between two nodes A and B are weighted by $Q_A^B$ and $Q_B^A$, i.e. what is known of the proportions of B's which are A's and of A's which are B's. Note that in terms of conditional probabilities we assume information on both P(A | B) and P(B | A), which contrasts with Bayesian networks (Pearl, 1988). Besides P(A) will be interpreted as P(A | X) where X stands for the set of all considered objects or in logical terms corresponds to the ever-true proposition. Hence all probabilities that we handle are (bounds of) conditional probabilities in a network where cycles are allowed, and no prior probability information is required in order to start the inference process in the approach described in this paper (contrary to Quinlan (1983)'s INFERNO system or Baldwin (1990)'s support logic programming).

Other works have been published, that handle probability bounds (see (Dubois et al., 1990) for a survey). However, these works always assume knowledge about unconditional probabilities (i.e. P(A) = P(A | X) in our framework) and are often oriented towards the computation of unconditional probabilities P(B). This is not true here. The reasoning systems of Bacchus (1990) aim at embedding the type of knowledge we deal with into a formal logical setting. Contrastedly our aim is to specify efficient inference algorithms.

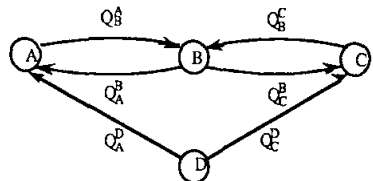

Figure 1 : an inference network

In the following sections, we are going to present computational methods that can handle imprecisely-known conditional probabilities. This work pursues an earlier investigation. In Dubois and Prade (1988), see also LéaSombé (1990), a first local pattern of reasoning, corresponding to the transitive chaining syllogism was studied. In (Dubois et al., 1990) two other local patterns enable us to estimate conditional probabilities involving conjunctions of events or contexts in their expression. A more complete set of propagation rules is presented in (Amarger et al. 1991).

After presenting the problem is section 2, section 3 recalls how our problem can be reduced to linear programming. Section 4 presents a generalized version of Bayes' theorem which can help improve the known bounds on an inference network in a single propagation step. Section 5 recalls the previously studied inference patterns involving conjunctions and disjunctions of two terms and discusses the handling of negation. Section 6 presents the general strategy that exploits two propagation rules in order to answer queries about conditional probabilities of interest. Section 7 illustrates the approach on an example. Section 8 discusses the handling of conjunction and disjunction in queries. Section 9 considers the introduction of independence assumptions in the chaining of conditional probabilities. In the conclusion, analogies with non-monotonic reasoning are pointed out.

## 2 STATEMENT OF THE PROBLEM

We suppose that we know some default rules containing numerical quantifiers or conditional probabilities such as "$Q_B^A$ A's are B's" or "if A then B with probability P(B | A)."

The objective of our research is to answer queries like : "what proportion of A's are C's ?," "what proportion of A's and B's are C's ?," "what proportion of C's are A's and B's ?," "what proportion of A's or B's are C's ?," or "what proportion of C's are A's or B's ? ;" or similar queries stated in terms of conditional probabilities, from the available knowledge about the values of other proportions or conditional probabilities. This corresponds respectively to evaluate the probabilities p = P(C | A), P(C | A ∩ B), P(A ∩ B | C), P(C | A ∪ B) or P(A ∪ B | C). The possible values of p usually form an interval $[p_*, p^*]$, and not just a single value, where $p_*$ is the lowest value and $p^*$ the highest value possibly taken by the conditional probability. Usually, a good local uncertainty propagation method will provide bounds that bracket $[p_*, p^*]$, i.e. the deduction method will be sound. If it supplies exact bounds, it is called complete. The inference patterns we shall use in the following sections are sound and are also complete when we consider just the elementary network corresponding to the statement of the pattern, i.e. they are said to be *locally complete*.

Our view of a knowledge base in this paper is thus a collection of general statements regarding a population X of objects ; these statements express in imprecise terms the proportions of objects in various subclasses of X, that belong to other subclasses. This knowledge base allows for answering queries about a given object, given a subclass to which it belongs (also called its "reference class" by Kyburg). To do so we just apply to this object the properties of this subclass, implicitly assuming that it is a typical element of this class. If more information become available for this object, we just change its reference class accordingly.

Computing bounds for P(B | A) is a matter of constraint propagation, and is not based on updating a probability

distribution, contrary to Bayesian reasoning. Namely, given that we know that an object of interest is in subclass A, we certainly do not interpret this fact as $P(A) = 1$. Indeed adding the constraint $P(A) = 1$ to the knowledge base may lead to modify the set of probability measures that obey the constraints induced by probability bounds (e.g. $P(C \mid A \cap B)$ must become equal to $P(C \mid B)$). This is because $P(A) = 1$ means that $X - A$ is an empty set ; on the contrary, a "fact" "$x \in A$" in our system just indicates that we look for the properties of members of subset A. Although in Bayesian reasoning, computing $P(B \mid A)$ is the same as assuming the posterior probability of A is 1 because the probability distribution on X is unique, these two operations no longer coincide with probability bounds : one is just focusing on a reference class (what can be said about B, for members of A?) while the other is knowledge updating (see Dubois and Prade, 1991b). Dubois and Prade (1991b) further discuss the difference between focusing and updating in the framework of belief functions. As for the difference between computing $P(B \mid A)$ and $P(B)$ when A is an accepted fact, this topic has been considered in the philosophical literature for a long time. See e.g. Suppes (1966).

## 3 A LINEAR PROGRAMMING METHOD

It has been shown in (Paass, 1988) that reasoning from numerically quantified general rules may be modelled as an optimization problem. Namely, if there are n atomic symbols in the network, there are $2^n$ possible worlds, and we can express all constraints on conditional probabilities as linear constraints where the variables $x_i$ correspond to the unknown probabilities of possible worlds i. The calculation of bounds on an unknown conditional probability $P(A \mid B)$ comes down to find the maximum and the minimum of a rational fraction whose numerator sums the probabilities $x_i$ of the possible worlds where A and B are true, and whose denominator sums the probabilities $x_i$ of the possible worlds where B is true, under the constraints induced by the already known probability bounds and to the requirement that the $x_i$'s sum to one.

The same approach may be used to solve any query. As we can see we are faced with a fractional linear programming problem of the form (P),

$$(P) \left\{ \text{Opti} \frac{c \cdot {}^t x}{d \cdot {}^t x}, x \geq 0 \text{ under } \mathbb{1} \cdot {}^t x = 1, M \cdot x \leq 0 \right\}$$

where $\mathbb{1}$ is the "unit vector," c, d, x are row vectors, M is a matrix, "Opti" is either "Max" or "Min," and t denotes the transposition, can be transformed into an equivalent linear program (P'). Indeed, as pointed out by (Charnes and Cooper, 1962), letting $y_i = x_i/(d \cdot {}^t x)$, we obtain :

$$(P') \left\{ \text{Opti } c \cdot {}^t y, y \geq 0 \text{ under } M \cdot y \leq 0, d \cdot {}^t y = 1 \right\}$$

So, as explained and exemplified in (Amarger et al., 1990), the calculation of bounds for $P(B \mid A)$ requires that two linear programs be solved (one to compute the exact lower bound and one to compute the exact upper bound). But, even if with this method we are able to precisely compute the best bounds bracketing the conditional probability of interest, it is hard to try to provide an explanation for the obtained results in terms of the available knowledge we start with.

This reduction of a fractional linear programming problem induced by probability constraints to a linear programming problem has been also pointed out and used in (van der Gaag, 1990), where also local computation methods are proposed on the basis of the decomposition of the linear system into subsystems, and exploiting independence relationships when they are known. Methods based on local inference patterns may provide less precise results (although they are guaranteed to be sound), but are faster and their results easier to explain.

## 4 GENERALIZED BAYES' THEOREM

In the framework of numerical quantifiers, because we manipulate conditional probabilities, it would be interesting to use the Bayes' theorem :

$$\forall A, B, \quad P(A \mid B) = P(B \mid A) \cdot P(A) / P(B)$$

But, in our approach we do not assume that $P(A)$ and $P(B)$ are known. A more general identity, where only conditional probabilities appear can be established :

Proposition 1 : Generalized Bayes' theorem

$$\forall A_1, \ldots, A_k, P(A_1 \mid A_k) = P(A_k \mid A_1) \prod_{i=1}^{k-1} \frac{P(A_i \mid A_{i+1})}{P(A_{i+1} \mid A_i)}$$

when all involved quantities are positive.

This identity is easily proved replacing conditional probabilities $P(A \mid B)$ by their expressions $P(A \cap B)/P(B)$. Note that this identity tells us that given a cycle $A_1, A_2, \ldots, A_k, A_{k+1} = A_1$ in a probabilistic network, the 2.k quantities $\{P(A_i \mid A_{i+1}), i \in ]k]\} \cup \{P(A_{i+1} \mid A_i), i \in ]k]\}$ (where $]k] \equiv ]0, k] \cap \mathbb{N}$) are not independent when positive: any 2.k - 1 of these quantities determine the remaining one. Now, because we use upper and lower probabilities, we extend this theorem as follows:

Proposition 2 : Generalized Bayes' theorem - upper/lower probabilities case.

Given k sets $A_1, A_2, \ldots, A_k$, with k > 2, the following inequalities should hold :

• lower bound :

$$\forall A_1, \ldots, A_k, P_*(A_1 \mid A_k) \geq P_*(A_k \mid A_1) \prod_{i=1}^{k-1} \underline{d}_{i, i+1}$$

with: $\forall i, j \in ]k], \underline{d}_{i,j} = P_*(A_i \mid A_j) / P^*(A_j \mid A_i)$.

• upper bound :

$$\forall A_1, \ldots, A_k, P^*(A_1 \mid A_k) \leq P^*(A_k \mid A_1) \prod_{i=1}^{k-1} \overline{d}_{i, i+1}$$

with: $\forall i, j \in ]k], \bar{d}_{i,j} = P^*(A_i|A_j)/P_*(A_j|A_i)$.

A simpler version of Proposition 2 is used by Fertig and Breese (1990) for arc reversal in influence diagrams where probabilities are incompletely known. Proposition 2 is the basis of a first inference rule for tightening probability bounds in a set of ill-known conditional probabilities.

Namely given a knowledge base $\mathcal{H} = \{(P_*(A_i | A_j), P^*(A_i | A_j)), i, j \in ]n]\}$ ; we can associate to it a network G with n nodes $A_1, A_2,..., A_n$ and whose arcs $(A_i, A_j)$ are weighted by $\underline{d}_{i,i+1}$. Proposition 2 leads to update $P_*(A | B)$, and $P^*(A | B)$ in one step as follows

$$P_*(A | B) = P_*(B | A) \cdot \max_{\substack{\text{over all paths } A_1, ..., A_k \text{ in G} \\ \text{with } 2 < k \le n, A_1 = A, A_k = B}} \left\{ \prod_{i=1}^{k-1} \underline{d}_{i,i+1} \right\} \quad (1)$$

$$P^*(A | B) = P^*(B | A) \cdot \min_{\substack{\text{over all paths } A_k, ..., A_1 \text{ in G} \\ \text{with } 2 < k \le n, A_1 = A, A_k = B}} \left\{ \prod_{i=1}^{k-1} 1/\underline{d}_{i,i+1} \right\} \quad (2)$$

The second update is easily explained noticing that $\bar{d}_{i,i+1} = 1/\underline{d}_{i+1,i}$. Note that these changes in probability bounds do not correspond to a revision of the knowledge, but only to constraint propagation steps ; namely the set of probability measures such that $\forall i, j, P_*(A_i | A_j) \le P(A_i | A_j) \le P^*(A_i | A_j)$ never changes.

As it can be guessed, the propagation of the constraint expressed by Proposition 1 is achieved by computing the longest (i.e. most weighted) elementary paths from A to B and from B to A in the network G where arcs (A,B) and (B,A) have been suppressed. Here the length of the path is the product of all weights of arcs in the path. For reason of computing accuracy, it is better to compute the length of the paths using a standard (max, +) path algebra, changing $\underline{d}_{i,i+1}$ into Log $\underline{d}_{i,i+1}$. Then any shortest path algorithm will do. Note that the length of a circuit $A_1,..., A_k, A_{k+1} = A_1$ in G is such that $\underline{d}_{1,2} \cdot \underline{d}_{2,3} \cdots \underline{d}_{k-1,k} \cdot \underline{d}_{k,1} \le 1$ ; indeed, this inequality reads

$$\prod_{i=1}^{k-1} P_*(A_i | A_{i+1}) \bigg/ \prod_{i=1}^{k-1} P^*(A_{i+1} | A_i) \le P^*(A_1 | A_k)/P_*(A_k | A_1)$$

and is a consequence of Proposition 1. Hence in the network with weights of the form Log $\underline{d}_{i,j}$, no circuit will be of positive length. Hence longest paths between nodes will always exist.

The constraint propagation steps (1) and (2) can be used as an inference rule that we shall denote BG (Bayes generalized) in the following.

## 5 LOCAL INFERENCE RULES

The first local inference pattern, already examined in (Dubois and Prade, 1988) and in (Dubois et al., 1990),
corresponds to the evaluation of a missing arc in the inference network, and can be viewed as the counterpart of node removal in influence diagrams.

The problem solved by this pattern, also called **"quantified syllogism rule"** (QS) is the following: given bounds on P(A | B), P(B | A), P(C | B) and P(B | C), what are the bounds on P(C | A) (thus removing node B). The following bounds can be shown to be the tightest ones :

lower bound :

$$P_*(C | A) = P_*(B | A) \max\left(0, 1 - \frac{1 - P_*(C | B)}{P_*(A | B)}\right) \quad (3)$$

upper bound :

$$P^*(C|A) = \min\left(1, 1 - P_*(B|A) + \frac{P_*(B|A) \cdot P^*(C|B)}{P_*(A|B)}\right),$$

$$\frac{P^*(B|A)P^*(C|B)}{P_*(A|B)P_*(B|C)}, \frac{P^*(B|A)P^*(C|B)}{P_*(A|B)P_*(B|C)}[1 - P_*(B|C)] + P^*(B|A)\right) (4)$$

The application of QS to the network with nodes A, B, C for the calculation of P(C | A) is denoted QS(C, B, A) = (C, A). For a proof that these bounds are optimal see (Dubois and Prade, 1988 ; Dubois et al., 1990).

This pattern can be extended to more than 3 nodes in sequence. It can be proved that optimality is preserved. Especially, given (A, B, C, D), it is equivalent to remove B first (computing P(C | A)), then C, or C first (computing P(D | B)), and then B, in order to get P(D | A), i.e. there is an associativity property.

Proposition 3 : QS(QS(D,C,B),A) = QS(D,QS(C,B, A))
  Proof : First, consider the network {A, B, C} ; applying QS we get bounds for P(C | A) and P(A | C). Then we could think of applying QS again in order to improve bounds of P(C | B) for instance. Clearly this process will not lead to improve these bounds. Indeed if these bounds were improved using P(B | A), P(A | B), and the calculated bounds on P(C | A), P(A | C), it would indicate that quantities P(B | C) or P(C | B) are related to P(B | A) or P(A | B). But this is clearly not true. Similarly the knowledge about P(D | C) and P(C | D) has no influence on P(B | C) and P(C | B), hence has no influence on the optimal bounds of P(C | A) and P(A | C). The optimality of the QS rule then implies that the result of applying it on P(C | A), P(A | C), P(C | D), P(D | C) will also give optimal bounds on P(D | A) and P(A | D). The same reasoning applies if we compute P(D | B), P(B | D) first. In both cases we get optimal bounds on P(D | A) and P(A | D). Associativity then follows from optimality. Q.E.D.

This result could also be derived by the study of the linear program associated to the network, looking for decomposability properties of the constraint matrix.

Clearly, this property of QS is very nice and easily generalizes to a network with any number of nodes. Thus on a "linear chain" beginning with node $A_1$ and ending with node $A_k$, we can apply QS iteratively, from left to right, in order to evaluate $P(A_k \mid A_1)$ for instance, without resorting to linear programming.

In (Dubois et al., 1990) the expression of bounds on $P(A \cap B \mid C)$ and $P(C \mid A \cap B)$ in terms of more elementary conditional probabilities $P(A \mid C)$, $P(C \mid A)$, $P(C \mid B)$, $P(B \mid C)$, $P(A \mid B)$ and $P(B \mid A)$ have been established starting with a complete network with nodes A, B, C. The case of disjunction is solved in (Amarger et al., 1991) where probabilities of the form $P(A \cup B \mid C)$ and $P(C \mid A \cup B)$ are explicitly obtained under the same setting. Disjunction and conjunction are addressed in Section 8 using the two rules QS and BG.

The case of negation is especially interesting. Indeed, given $P(B \mid A)$ and $P(A \mid B)$, we obviously know $P(\neg B \mid A)$ and $P(\neg A \mid B)$ where $\neg$ denotes complementation. However it is easy to verify that $P(A \mid \neg B)$ and $P(B \mid \neg A)$ remain totally unknown. It is indeed easy to check that

$$P(A \mid \neg B) = P(A \mid B) \cdot \left((1/P(B \mid A)) - 1\right) \cdot P(B)/P(\neg B) \quad (5)$$

In other words, answering queries of the form "How many not B's are A's" require the knowledge of unconditional probabilities. A possible other way of dealing with negation is to introduce the closed world assumption which can be stated as follows : if sets A, B and $C_i$, $i \in \,]n]$ appear in the network, then let us assume that the universe is reduced to $A \cup B \cup \bigcup_{i \in \,]n]} C_i$. In other words, we assume that the set $\neg A \cap \neg B \cap \bigcap_{i \in \,]n]} \neg C_i$ is empty, or at least that $P(\neg A \cap \neg B \cap \bigcap_{i \in \,]n]} \neg C_i) = 0$. In the trivial case where we consider the classes A and B only, it leads to $P(\neg A \cap \neg B) = 0$, and then $P(A \mid \neg B) = P(A \mid A \cap \neg B) = 1$. So, if we "open" the world by considering C also, then we assume $P(\neg A \cap \neg B \cap \neg C) = 0$. Since $\neg B = [\neg B \cap (A \cup C)] \cup [\neg B \cap \neg(A \cup C)] = \neg B \cap (A \cup C) \cup (\neg A \cap \neg B \cap \neg C)$, then $P(\neg B) = P(\neg B \cap (A \cup C))$. Thus we change the question "what is the value of $P(A \mid \neg B)$ ?" into "what is the value of $P(A \mid \neg B \cap (A \cup C))$ ?". A systematic way of dealing with these questions require a proper handling of Boolean expressions in conditional probabilities.

## 6 A CONSTRAINT PROPAGATION BASED ON INFERENCE RULES

In the previous sections, we have presented two local inference rules, and now, the problem is to use these rules in order to perform automated reasoning with the whole network. The aim of this section is to build a reasoning strategy in order to be able to answer any *simple* query (i.e. a query of the form "what is the proportion of A's which are C's?," where A and C are atoms in the language). The network is supposed to be made out of simple conditional probabilities of the form $P(A \mid B)$ where A and B are atomic symbols.

Graphically, to answer a query like "what proportion of X's are Y's ?" is equivalent to generate the new arc $\langle X, Y \rangle$ in a network like the one of Figure 1. Our approach is local in the sense that the patterns are designed to provide answers to particular queries using local inference rules. Consequently, one can observe the influence of each piece of knowledge on the result ; global methods do not offer such a possibility. Even though a particular pattern corresponds to an elementary network, the inference patterns can work on any network, whatever its structure, unlike the Bayesian approach which needs an acyclic network topology (e.g. directed cycles are prohibited) adapted to the propagation mechanism ; see (Lauritzen and Spiegelhalter, 1988 and Pearl, 1988).

Of course, in practice, in order to answer a particular query, it may exist several possibilities for applying the inference patterns to the network, corresponding to different paths. Since the inference rules are sound, one can easily combine the different results provided by all the applications of rules because their intersection still provides a sound result. Indeed, let us suppose that $Q_1 = [p_{*1}, p^*_1]$ and $Q_2 = [p_{*2}, p^*_2]$ are two intervals that contains the value p we want to estimate ; then we have : $p \in Q_1 \cap Q_2 = [\max(p_{*1}, p_{*2}), \min(p^*_1, p^*_2)]$. This generalizes to the intersection of any number of intervals ; and the emptiness of the intersection would be the proof that the data we start with are not consistent.

We will first use a saturation strategy in order to extract as much information as we can from the network, namely, try to get probability intervals as tight as possible for all conditional probabilities $P(A \mid B)$. The result is called the *saturated network*.

We are going to use two tools : rule QS (corresponding to the basic quantified syllogism) presented in Section 5., in order to add links to the network, and the generalized Bayes' theorem (rule BG) presented in Section 4.

Step 1 : recursively apply QS, to generate the missing arcs. This step is performed until the probability intervals can no more be improved.
Step 2 : recursively apply BG to improve the arcs generated by Step 1.

Then, the general algorithm is :
(a)  perform Step 1
(b)  perform Step 2
(c)  if the probability intervals have been improved go to (a), otherwise stop

Note that the two steps are very complementary. Indeed, step 1 uses an optimal rule but a local one, while step 2 uses a suboptimal method but considers more than 3-tuples of nodes.

Another important problem encountered in inference system is the consistency of the knowledge base. Using the global method presented in Section 3., if one of the two linear programs we have to solve (or both) has no solution, we can say that there is an inconsistency in the constraints of the linear programs, i.e. an inconsistency in the knowledge base. Solving only one linear program is enough to find out an inconsistency (if any) among the constraints expressing the knowledge base. If there is some inconsistency, exhibiting the Simplex array, we will be able to determine where is the inconsistency, i.e. which arcs are inconsistent. So, our system is of the following general form :
(a) consistency checking by linear programming
if an inconsistency is detected, exit
(b) saturation of the network
(c) answering user's queries.
The considered queries are of the form P(A | B) ?.

Using results in established in (Dubois et al., 1990; Amarger et al., 1991) we can also handle queries of the form P(A ∪ B | C) ?, P(A ∩ B | C) ?, P(C | A ∩ B) ?,...Of course, steps (a) and (b) may take a long time computation, but they only are performed once for all at the beginning of the session, in order to ensure that the user works with a consistent knowledge base, and to make all the information explicit.

## 7 AN EXAMPLE

In this section, our purpose is to point out the results given by both the quantified syllogism and generalized Bayes' theorem. The algorithm we use is written in "C" on a Sun 3/50 workstation without arithmetical co-processor and the Floyd algorithm is used to compute the longest paths (see (Gondran and Minoux, 1985) for instance). The example we use is already considered in (Dubois et al., 1990), and is pictured in Figure 2 and, in the following, we use the incidence matrix notation to let the saturated network be more readable.

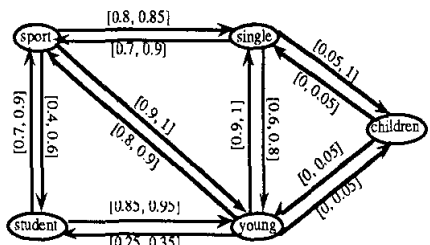

Figure 2

So, using the above algorithm (the details are given in (Amarger, Dubois, Prade 1991)), we get the "saturated" network (the improved bounds are underlined):

|          | student      | sport        | single       | young        | children     |
|----------|--------------|--------------|--------------|--------------|--------------|
| student  | [1.00;1.00]  | [0.90;0.90]  | [0.61;1.00]  | [0.85;0.85]  | [0.00;0.27]  |
| sport    | [0.40;0.40]  | [1.00;1.00]  | [0.85;0.85]  | [0.90;0.96]  | [0.00;0.15]  |
| single   | [0.22;0.36]  | [0.70;0.70]  | [1.00;1.00]  | [0.80;0.80]  | [0.05;0.10]  |
| young    | [0.35;0.35]  | [0.84;0.88]  | [0.90;0.90]  | [1.00;1.00]  | [0.00;0.05]  |
| children | [0.00;0.09]  | [0.00;0.13]  | [0.00;0.05]  | [0.00;0.04]  | [1.00;1.00]  |

The computation of the complete "saturated" matrix was made in 10 seconds (CPU and I/O time).

The optimal solution computed by the global method presented in Section 3., and in (Amarger et al., 1990) is exactly the same as the one computed by the "local method" based on QS and BG. Let us note that the "global method" is written in "C", on a Sun 3/50 workstation, without arithmetical co-processor; and the computation of each element of the "optimal" matrix is made in 12 seconds (CPU and I/O time). So, combining a locally optimal method (QS) with a global but suboptimal method (generalized Bayes' theorem), we get results as good as the ones given by a globally optimal method (Simplex based method of Section 3.), but with a much smaller computation time, in our example.

## 8 CONJUNCTION AND DISJUNCTION

The results involving conjunction and disjunction solved in previous papers are not general enough to be very useful in practice. Their merits are but tutorial. Especially, their extension to disjunctions and conjunctions of more than two terms look untractable in an analytic form. Even the case when only three symbols A, B and C are involved, and where bounds on the six conditional probability values involving these symbols are known, will lead to unwieldy expressions because the six values are related via the generalized Bayes' theorem.

A more realistic approach to the problem of handling disjunctions and conjunctions is to introduce new nodes in the network, that account for the concerned conjunctions and disjunctions, and apply the iterative algorithm (or linear programming) to answer the query. As an example, let us consider the query "what is the probability of C given A and B", where the background network includes nodes A, B, C only; (see Figure 3)

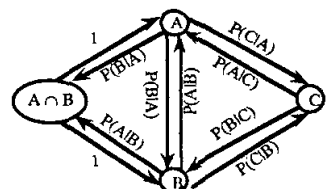

Figure 3 : Introducing a new node "A ∩ B"

To deal with this problem we create a node named A ∩ B. A description of the conjunction in terms of conditional probabilities leads to force $P_*(A | A \cap B) = 1$, $P_*(B | A \cap B) = 1$, $P(A \cap B | B) = P(A | B)$ and $P(A \cap B | A) = P(B | A)$, and to add these arcs to the network (see Figure 3). Then the calculation of P(C | A ∩ B) can be addressed by the repeated use of the Quantified Syllogism pattern and the generalized Bayes rule in this network.

In order to catch a feeling of what kinds of results can be produced by this method, let us deal with the case when the six values P(A | C), P(C | A), P(B | C), P(C | B), P(A | B), P(B | A) are precisely known in Figure 3. Of course they obey the generalized Bayes theorem, so that only five of them need to be known. The calculation of bounds for P(C | A ∩ B) can be performed by applying twice the syllogism rule, cancelling A between A ∩ B and C, and cancelling B between A ∩ B and C. Applying (1) and (3) with the following substitution : A becomes A ∩ B, B becomes A, we get

$$\max\left(0, 1 - \frac{1 - P(C | A)}{P(B | A)}\right) \le P(C | A \cap B) \le \min\left(1, \frac{P(C | A)}{P(B | A)}\right)$$

Similarly, exchanging A and B in the above inequalities, we get :

$$m \times \left(0, 1 - \frac{1 - P(C | B)}{P(A | B)}\right) \le P(C | A \cap B) \le \min\left(1, \frac{P(C | B)}{P(A | B)}\right)$$

Joining these results together, we obtain

$$\max\left(0, 1 - \frac{1 - P(C | A)}{P(B | A)}, 1 - \frac{1 - P(C | B)}{P(A | B)}\right) \le P(C | A \cap B) \quad (6)$$

$$P(C | A \cap B) \le \min\left(1, \frac{P(C | B)}{P(A | B)}, \frac{P(C | A)}{P(B | A)}\right) \quad (7)$$

It can be checked that this is exactly what has been obtained in (Dubois et al., 1990), i.e. when we have no knowledge about P(B | C) and P(A | C). To improve these bounds requires the use of the generalized Bayes theorem. As shown in (Dubois et al., 1990) only the lower bound of P(C | A ∩ B) can be improved knowing P(B | C) and P(A | C). However the following extra inequalities are not related to the generalized Bayes theorem nor to the quantified syllogisms

$$P(C | A \cap B) \ge \frac{P(C | A)}{P(B | A)} + \frac{P(C | B)}{P(A | B)} \cdot \left(1 - \frac{1}{P(B | C)}\right) \quad (8)$$

$$P(C | A \cap B) \ge \frac{P(C | B)}{P(A | B)} + \frac{P(C | A)}{P(B | A)} \cdot \left(1 - \frac{1}{P(A | C)}\right) \quad (9)$$

These inequalities are simple consequences of the additivity of probabilities applied to A ∩ B ∩ C under the form

P(A ∩ B ∩ C) = P(A ∩ C) + P(B ∩ C) - P((A ∩ C) ∪ (B ∩ C))
≥ P(A ∩ C) + P(B ∩ C) - P(C)

Hence additivity is not presupposed by the description of node A ∩ B in Figure 3. Proceeding similarly for P(A ∩ B | C), the syllogism rule leads to the following bounds

$$m \times \left(0, P(A | C)\left(1 + \frac{(P(B | A) - 1)}{P(C | A)}\right), P(B | C)\left(1 + \frac{(P(A | B) - 1)}{P(C | B)}\right)\right)$$
$$\le P(A \cap B | C) \le$$
$$\min\left(P(A | C), P(B | C), \frac{P(A | C)P(B | A)}{P(C | A)}, \frac{P(B | C)P(A | B)}{P(C | B)}\right)$$

Note that in the above expression, the two last terms in the 'min' are equal due to the generalized Bayes' theorem. Using results in (Dubois et al., 1990), it can be checked that the upper bound is optimal while the lower bound is sound but not optimal. Indeed we do not recover the obvious bound, again related to additivity :

P(A ∩ B | C) ≥ max(0, P(A | C) + P(B | C) - 1) (10)

More specifically, given only P(A | C) = 1 and P(B | C) = 1, the repeated use of the syllogism rule and the generalized Bayes' rule are not capable of producing P(A ∩ B | C) = 1 (a result produced by the above bound). Indeed, if we add the node AB to represent A ∩ B, we have to saturate the following network

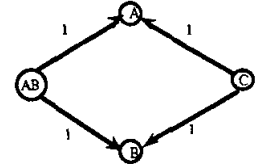

Figure 4

All that this network tells is that AB ⊆ A ∩ B and C ⊆ A ∩ B, but clearly, AB ∩ C can be anything. Also, even assuming that P(A | B) ≠ 1 and P(B | A) ≠ 1 are known and letting P(AB | A) = P(B | A), P(AB | B) = P(A | B) cannot improve the lower bound of P(AB | C) using the syllogism rule, nor the generalized Bayes rule. This point indicates that some of the lower bounds already obtained in (Dubois et al., 1990), for the conjunction will be useful to implement, in order to improve the performance of the iterative procedure, i.e. the inequalities (8), (9) and (10).

Another point to notice is that the constraint P(AB | A) = P(B | A) is stronger than letting P*(AB | A) = P*(B | A), $P_*(AB | A) = P_*(B | A)$, when only bounds on P(B | A) are known, indeed, the equality of the bounds can go along with the inequality P(AB | A) ≠ P(B | A). Let us consider the query about P(C | A ∪ B). To deal with this case, we create a node named A ∪ B, and arcs joining this node to the network, so as to describe the disjunction in terms of conditional probabilities namely P(A ∪ B | A) = 1 and P(A ∪ B | B) = 1. The calculation of P(A | A ∪ B) and P(B | A ∪ B) is slightly less straightforward, namely

$$P(A | A \cup B) = \frac{P(A)}{P(A \cup B)} = \frac{P(A)}{P(A) + P(B) - P(A \cap B)}$$
$$= \frac{P(A | B)}{P(A | B) + P(B | A) - P(A | B) \cdot P(B | A)}$$

since P(B)/P(A) = P(B | A)/P(A | B). The complete study of this case is left to the reader. A lack of optimality similar to the one encountered with conjunction will be observed.

## 9 INDEPENDENCE ASSUMPTIONS

Although our approach does not require independence assumptions, it should be possible to use them if they hold, in order to improve bounds. This section gives preliminary results on that point, for the syllogism rule

QS. Let us consider conditional independence relations. There are three possible ones on $\{A, B, C\}$:

i)   $P(B \cap C \mid A) = P(B \mid A) \cdot P(C \mid A)$
ii)  $P(A \cap C \mid B) = P(A \mid B) \cdot P(C \mid B)$
iii) $P(A \cap B \mid C) = P(A \mid C) \cdot P(B \mid C)$

First, note that i) and iii) are symmetric with respect to each other, exchanging C and A. We shall thus just consider i) and ii). ii) has already been considered in the introduction and we shall check that we cannot do better: ii) is indeed equivalent to the irrelevance property $P(C \mid B) = P(C \mid A \cap B)$. Hence the independence property can be exploited by substituting $P(C \mid B) = P(C \mid A \cap B)$ in the bounds on $P(C \mid A \cap B)$ (equations (6), (7)). Only the bounds where $P(C \mid A)$ appear are useful. We get (for precise values)
$1 - (1 - P(C \mid A)/P(B \mid A)) \leq P(C \mid B) \leq P(C \mid A)/P(B \mid A)$
from which it follows :
$P(C \mid B) \cdot P(B \mid A) \leq P(C \mid A) \leq 1 - P(B \mid A) + P(C \mid B) \cdot P(B \mid A)$ (11)
the lower bound improves (3) and the upper bound improves the second term in the general upper bound (4). Particularly, when $P(B \mid A) = 1$ it can be checked that ii) entails $P(C \mid A) = P(C \mid B)$. For bounds on $P(A \mid C)$, just exchange A and C in the above inequalities, and get
$P(A \mid B) P(B \mid C) \leq P(A \mid C) \leq 1 - P(B \mid C) + P(A \mid B) P(B \mid C)$ (12)
The above inequalities can influence $P(C \mid A)$ using the generalized Bayes rule since
$P(A \mid C) = P(C \mid A) \cdot P(A \mid B) \cdot P(B \mid C) / P(B \mid A) \cdot P(C \mid B)$
can be substituted in (12) and enable to catch the inequality
$P(C \mid A) \leq \dfrac{P(B \mid A) P(C \mid B)}{P(A \mid B) P(B \mid C)} \cdot (1 - P(B \mid C) + P(A \mid B) P(B \mid C))$ (13)
that comes on top of (11) (the lower bound of (11) is obtained again this way). It improves the last term appearing in the upper bound in (4).

Let us turn to i). It yields a new expression for $P(C \mid A)$ under the from $P(B \cap C \mid A)/P(B \mid A)$. Let us write it by letting $P(A \mid B \cap C)$ appear; using the generalized Bayes rule:
$$P(C \mid A) = \dfrac{P(A \mid B \cap C) \cdot P(C \mid B)}{P(A \mid B)}$$
Now using optimal bounds (6) and (7) on $P(A \mid B \cap C)$, and given that $P(A \mid C)$ is unknown there comes
$\max\left(0, 1 - \dfrac{1 - P(C \mid B)}{P(A \mid B)}\right) \leq P(C \mid A) \leq \min\left(1, \dfrac{P(C \mid B)}{P(A \mid B)}\right)$ (14)

Again, if $P(C \mid B) = 1$, we conclude that $P(C \mid A) = 1$. Moreover if $P(A \mid B) = 1$, then $P(C \mid A) = P(C \mid B)$. The lower bound in (14) improves (3), and the upper bound may improve the third term in (4).
Independence assumption iii) leads to a similar bracketting of $P(A \mid C)$, just exchanging C and A in (14) :
$\max\left(0, 1 - \dfrac{1 - P(A \mid B)}{P(C \mid B)}\right) \leq P(A \mid C) \leq \min\left(1, \dfrac{P(A \mid B)}{P(C \mid B)}\right)$ (15)
(15) and the generalized Bayes rule enable special bounds for $P(C \mid A)$ to be found under assumption iii), namely :

$\dfrac{P(B \mid A)}{P(B \mid C)}\left[1 - \dfrac{1 - P(C \mid B)}{P(A \mid B)}\right] \leq P(C \mid A) \leq \dfrac{P(B \mid A)}{P(B \mid C)}$ (16)

Again the lower bound in (16) improves (3), and the upper bound may improve the third term in (4).

To summarize, when independence assumptions are declared, namely i), ii), iii), bounds on $P(C \mid A)$ given in (3) and (4) can be improved by means of (14), (11) and (13), and (16) respectively. Of course, these types of independence assumption can be more directly exploited in queries involving conjunctions or disjunctions.

## 10 CONCLUSION

The approach proposed in this paper to handle conditional probabilities in knowledge networks presupposes assumptions that contrast with the ones underlying Bayesian networks. In Bayesian networks, a single joint probability distribution is reconstructed from the acyclic network using conditional independence assumptions, and given some a priori probabilities on the roots of the acyclic network. Here, nothing is assumed about a priori (unconditional) probabilities, no independence assumption is taken for granted, and, the more cycles there are, the more informative the network is.

Results obtained so far indicate that the two inference rules that we use in turn, namely the syllogism rule (QS) and the generalized Bayes' theorem (BG), are powerful and can compete with a brute force linear programming approach, as regards the quality of the obtained probability bounds. Our inference technique seems to be more efficient than linear programming since each run of each step of the inference procedure is polynomial in the number of nodes in the network. However, more investigation is needed on complexity aspects, and to better grasp the distance to optimality of the inference procedure.

It has been indicated how to deal with conditional probabilities involving conjunctions and disjunctions of two terms, and negation of terms. However the obtained optimal bounds are rather heavy mathematical expressions for conjunctions and disjunctions, and it seems difficult to extrapolate them to more than two terms. It has been shown how to solve the problem of conjunction and disjunction by introducing auxiliary nodes in the original network. In the future, we plan to treat negation likewise and to generalize the node addition approach to the combination of more than two primitive terms.

In the long run, we plan to develop a computerized tool (parts of which are already implemented) that can handle a knowledge base in the form of a pair $(W, \Delta)$ where $W$ is a set of facts and $\Delta$ a sets of conditional probabilities. A query Q can then be solved by computing $P(Q \mid W)$ where $W$ is the conjunction of available facts, and $P(Q \mid W)$ is obtained under the form of bounds derived from the saturated network built with Δ. This mode of reasoning is similar to what happens in non-monotonic logic. More specifically some of the propagation rules proposed here bear some interesting analogies with some derived inference rules in a well-behaved non-monotonic logic. For instance the BG rule corresponds to

$$\frac{\alpha_1 \mathrel{\vdash\!\sim} \alpha_2, \alpha_2 \mathrel{\vdash\!\sim} \alpha_3, \ldots, \alpha_{n-1} \mathrel{\vdash\!\sim} \alpha_n, \alpha_n \mathrel{\vdash\!\sim} \alpha_1}{\alpha_1 \mathrel{\vdash\!\sim} \alpha_n} \text{ (loop)}$$

where $\mathrel{\vdash\!\sim}$ denotes the non-monotonic consequence relation discussed in (Kraus and al, 1990). The QS rule gives

$$\frac{\alpha_1 \mathrel{\vdash\!\sim} \alpha_2, \alpha_2 \mathrel{\vdash\!\sim} \alpha_1, \alpha_2 \mathrel{\vdash\!\sim} \alpha_3}{\alpha_1 \mathrel{\vdash\!\sim} \alpha_3} \text{ (equivalence)}$$

the basic lower bound for $P(A \cap B \mid C)$ (see (Amarger et al., 1990)) corresponds to

$$\frac{\gamma \mathrel{\vdash\!\sim} \alpha, \gamma \mathrel{\vdash\!\sim} \beta}{\gamma \mathrel{\vdash\!\sim} \alpha \wedge \beta} \text{ (right and)}$$

These analogies are no longer surprizing since such kinds of links between probabilistic reasoning and non-monotonic logic have been already laid bare by (Pearl, 1988) and the authors (Dubois and Prade, 1991). But the correspondence pointed out above suggests to consider a nonmonotonic logic where primitive inference rules are the above rules, i.e.rules which are usually considered as derived ones. This point is worth studying in the future.

Among topics of interest for future research, a more detailed comparison with the Bayesian approach would be quite interesting, of course. It would allow the loss of information due to the absence of a priori probabilities to be quantified. It has been demonstrated how to allow for independence assumptions in our approach. Clearly it generates non-linear constraints in the optimization problem associated to a query. But it seems that the inference procedure can cope with these assumptions in a nicer way, just by modifying the constraint propagation rules accordingly. Another topic is the extension of our method to fuzzy quantifiers, already considered in (Dubois and Prade, 1988) for the syllogism rule.

**Acknowledgements**
This work is partially supported by the DRUMS project (Defeasible Reasoning and Uncertainty Management Systems), funded by the Commission of the European Communities under the ESPRIT BRA n° 3085.